\documentclass{article}

\usepackage{graphicx}
\usepackage{booktabs, multirow, siunitx}

\usepackage[main, final]{neurips_2026}

\usepackage[utf8]{inputenc} 
\usepackage[T1]{fontenc}    
\usepackage{hyperref}       
\usepackage{url}            
\usepackage{booktabs}       
\usepackage{amsfonts}       
\usepackage{nicefrac}       
\usepackage{microtype}      
\usepackage{xcolor}         

\title{Effect-Driven Skill Abstractions for~Offline~Reinforcement~Learning}

\author{%
  Burcu Kilic\textsuperscript{1},
  David Drexel\textsuperscript{2},
  Emre Ugur\textsuperscript{1,3},
  Justus Piater\textsuperscript{2,4} \\[6pt]
  \textsuperscript{1}Department of Computer Engineering, Bogazici University, Türkiye \\
  \textsuperscript{2}Department of Computer Science, University of Innsbruck, Austria \\
  \textsuperscript{3}Department of Computer Engineering, Middle East Technical University, Türkiye \\
  \textsuperscript{4}Digital Science Center, University of Innsbruck, Austria \\[4pt]
  \texttt{burcu.kilic@std.bogazici.edu.tr}
}

\begin{document}

\maketitle

\begin{abstract}
 We seek to learn discrete skill abstractions to facilitate long-horizon, goal-conditioned reinforcement-learning (RL) tasks. Recent approaches have shown success by quantizing actions via clustering or reconstruction. However, these objectives lead to skills that are not directly related to what the current state affords or what the goal requires. Instead, we argue that goal-directed skills should be driven by \emph{how they affect the state} of the agent. To realize this, we introduce a novel quantization method for action chunks that optimizes next-state predictions conditioned on the current state and action. This objective encourages skills to be distinguished by the future states they lead to, resulting in a vocabulary that spans diverse outcomes reachable from a given state. Our framework can easily be paired with common offline RL methods by replacing their continuous actor head with a discrete one that selects among the learned skills. We can execute these skills in the environment by converting them back to continuous actions using a trained low-level controller. Experiments on benchmark navigation and manipulation tasks largely show substantial and consistent improvements over both the continuous-actor baselines and reconstruction-based discretizations. Our project page can be found at: \url{https://burcukilic.github.io/projects/effect-driven-skill-abstractions/}
\end{abstract}

\section{Introduction}

Offline Reinforcement Learning (RL) aims to learn decision-making policies from collected interaction data to eliminate the safety and sample-efficiency issues of online exploration~\cite{levine2020offlinerl}. This paradigm is especially useful in goal-conditioned settings~\cite{kaelbling1993Learning, park2025ogbench, schaul2015universal} where a single policy must reach various targets in the environment, making online exploration unsafe and expensive. However, training stable goal-conditioned policies remains challenging for long-horizon tasks in sparse-reward settings, especially in high-dimensional continuous action spaces, where the policy may output actions outside the training dataset's support, leading to unstable training and unreliable value estimates. 

One way to mitigate this is to convert the continuous problem into a discrete one. It is generally more straightforward to optimize a policy over a finite set of actions than over an uncountable set~\cite {dadashi2022continuous}. The simplest approach to discretizing a continuous action space is to uniformly partition each dimension into discrete bins. However, such a partition would suffer from the curse of dimensionality in high-dimensional spaces. A better alternative is to learn the discretizations from data. AQuaDem~\cite{dadashi2022continuous} and SAQ~\cite{luo2023action} learn state-conditioned action discretizations, improving both training stability and task performance over continuous baselines. These methods discretize actions at the level of a single timestep. However, for long-horizon tasks, behavior should be temporally extended as a prerequisite for tractable learning and reasoning~\cite{sutton1999between, konidaris2019necessity}. Recent studies quantize \emph{action chunks} rather than single-step actions to produce discrete skills~\cite{pinto2022bet, lee2024vqbet, zheng2024prise, mete2024quest}. These methods perform quantization by reconstructing or clustering actions, which is a sensible objective to capture the multimodality of the demonstrator's behavior.

We argue that for goal-conditioned tasks, what makes a learned skill useful is the \emph{effect} it has on the state, not the particular sequence of low-level control commands that produces it. In this respect, in the unsupervised skill discovery literature, DIAYN ~\cite{eysenbach2019diversity} learns diverse skills by making the states they visit distinguishable, and DADS~\cite{sharma2020dads} trains skills whose next state, given the current state, is predictable. However, neither approach performs explicit discretization; they observe diverse behavioral patterns in the data but do not represent them as separate entities or leverage them as reusable primitives. Inspired by this consequence-driven perspective, we quantize action chunks based on next-state predictions, grounding each discrete skill in its effect on the state rather than in its low-level execution. Conditioning this effect prediction mechanism on the current state further ensures that each discrete skill corresponds to an outcome that the state already offers in the dataset.
In this sense, the resulting skill codes behave as state-dependent \emph{affordances}~\cite{gibson1977theory, khetarpal2020can}, which are skill possibilities available to the agent in the current state provided by the environment. Through this quantization framework, we aim to discover a discrete vocabulary of skills that yield meaningful outcomes, i.e., outcomes that allow the system to effectively sequence these skills to achieve given goals.

In summary, we introduce a novel action chunk quantization method that discovers goal-driven discrete skills by predicting the outcomes they generate. Our system also learns a separate low-level controller network to translate the discrete skills back to continuous actions. Both the skill vocabulary and the controller are policy-independent, meaning that they can be paired with various offline RL methods. To demonstrate policy independence, we derive discrete variants of four standard offline RL methods, which are trained as high-level policies that output discrete skills instead of continuous actions. We evaluate our method across state- and pixel-based navigation and manipulation domains, including Maze, Puzzle, and visual tasks from the Offline Goal-Conditioned RL Benchmark (\textbf{OGBench})~\cite{park2025ogbench}, Push-T dataset~\cite{chi2023diffusion}, and D4RL Kitchen~\cite{fu2021d4rl} against both continuous baseline policies and reconstruction-based or single-step action quantization alternatives, demonstrating that our method consistently outperforms all baselines.

\begin{figure}
  \centering
  \includegraphics[width=\linewidth]{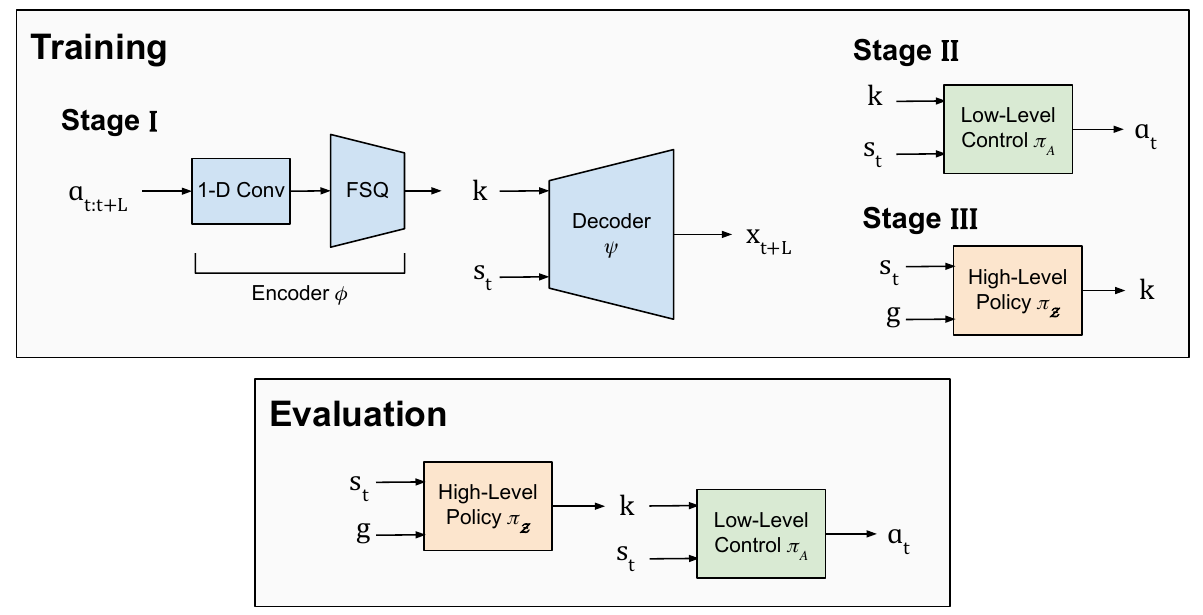}
  \caption{Three-stage training and evaluation overview of our proposed framework. \textbf{Stage I}: Action chunk quantization, optimized for predicting next-state features. 
  \textbf{Stage~II}: Learning to execute the skill codes. \textbf{Stage III}: Training a goal-conditioned discrete policy that produces skill codes. During evaluation, the high-level policy $\pi_{\mathcal{Z}}$ produces discrete skill codes that are converted to continuous actions by the trained low-level controller and are executed in the environment.}
  \label{fig:stages}
\end{figure}

\section{Related Work}

\textbf{Goal Conditioned Offline Reinforcement Learning}.
Offline RL~\cite{levine2020offlinerl} aims to learn a policy entirely from a given dataset of transitions without requiring further environment interactions. Several methodologies have been proposed, such as simple supervised learning of actions~\cite{ghosh2021learning}, implicit Q-learning~\cite{kostrikov2021offline} to avoid the difficulty of estimating unseen actions, and contrastive learning~\cite{eysenbach2022contrastive} to mitigate out-of-distribution action generation. In this paper, we focus on goal-conditioned offline RL for multi-goal sparse-reward tasks~\cite{park2025ogbench, yang2022rethinking}. Early successful approaches used conditioning with hindsight relabeled goals~\cite{andrychowicz2017hindsight, li2020generalized}. Quasimetric reinforcement learning (QRL)~\cite{wang2023optimal} proposed to use quasimetric models to learn the goal-conditioned value function. One family of approaches tackles long-horizon tasks by hierarchical learning of high-level decisions and low-level executions. In this context, HIQL~\cite{park2023hiql} constructs subgoal states in the high-level policy and navigates to subgoals in the low-level policy. Recently, OTA~\cite{ahn2025optionaware} used temporally abstracted value learning to enable robustness in long-horizon tasks. Our work focuses on a similar high-level temporally abstracted policy, but we differ in learning policy-independent abstractions for skills.

\textbf{Discretized Skills}.
Recent work proposed various discretization strategies for continuous action spaces, which have been shown to boost the performance of on-policy learning~\cite{tang2020discretizing}. AQUaDem~\cite{dadashi2022continuous} learned state-conditioned, single-step discrete actions by training with a soft-minimum variant of the Behavior Cloning objective, and then used these discrete actions to train an online RL policy. 
SAQ~\cite{luo2023action} used a vector quantization (VQ) codebook~\cite{van2017neural} to quantize single-step actions, trained discrete offline-RL policies, and showed improved performance over continuous versions.
These methods focus on the discretization of the action space; however, in this paper, we are interested in utilizing the temporal structure of actions to find skill abstractions for action chunks. 
In this context,
Behavior Transformer (BeT)~\cite{pinto2022bet} was introduced as a two-stage learning process for first clustering action chunks with k-means and then learning a transformer-based policy that predicts action codes. 
VQ-BeT~\cite{lee2024vqbet} improved this architecture by using a residual VQ codebook to quantize action chunks. Although these methods showed significant improvement in capturing the multimodality of action trajectories, they performed discretization by pure reconstruction and didn't investigate capturing the available action possibilities that a particular state can offer.
Primitive Sequence Encoding (PRISE)~\cite{zheng2024prise} used sequence compression to learn temporal action abstractions and leveraged next-state predictions in the process. TAP~\cite{jiang2023efficient} proposed efficient planning in the latent action space. Similarly to our method, Quantized Skill Transformer(QueST)~\cite{mete2024quest} used Finite Scalar Quantization (FSQ)~\cite{mentzer2024finite} to obtain temporal skill abstractions and trained an offline discrete policy, but used a reconstruction objective. In our method, we discover goal-driven skill abstractions by discretizing action chunks based on effect predictions and show that the discovered skills are compatible with different kinds of offline RL policies. In other words, our work defines reusable temporal skills directly by \textit{what they enable the agent to achieve}.

\section{Preliminaries}

\subsection{Problem Setting: Goal-Conditioned Reinforcement Learning}
Goal-conditioned reinforcement learning (GCRL) is formulated as a Markov decision process (MDP) $\mathcal{M} = (\mathcal{S}, \mathcal{A}, \mathcal{G}, r, \mu, p, \gamma)$ extended with goals. Here, $\mathcal{S}$ is the state space, $\mathcal{A}$ is the action space, $\mathcal{G}$ is the goal space that we assume to be equivalent to the state space ($\mathcal{G} \equiv \mathcal{S}$), $r(s, g, a)$ is the reward function, $\mu(s_0)$ is the initial state distribution, $p(s'\mid s, a)$ is the transition function, and $\gamma$ is the discount factor. 
Unlike online RL, offline RL cannot interact with the environment. Instead, it is provided with a transition dataset $\mathcal{D} = \{(s_t, a_t, r_{t+1}, s_{t+1})\}$. 
GCRL aims to learn an optimal goal-conditioned policy $\pi(a \mid s,g)$ from $\mathcal{D}$ that maximizes the objective
\begin{equation}
    \mathbb{E}_{g \sim p_\mathcal{G}(\cdot),\, \tau \sim p^\pi(\tau \mid g)}
    \left[ \sum_{t=0}^{\infty} \gamma^t\, r(s_t, g, a_t) \right]
\end{equation}
where $p_\mathcal{G}$ is the distribution of goals and $p^\pi(\tau\mid g)$ defines the trajectory distribution induced by the policy $\pi$ conditioned on goal $g$. 
In this study, we work with variants of some well-known GCRL methods. Goal-conditioned behavior cloning (\textbf{GCBC})~\cite{ghosh2021learning} is a simple imitation learning technique that learns actions from given states and goals in the dataset. Goal-Conditioned Implicit Q-Learning (\textbf{GCIQL}) is a goal-conditioned variant of IQL~\cite{kostrikov2021offline} that learns value functions V and Q using expectile regression~\cite{newey1987asymmetric}. \textbf{GCIVL} is a variant of GCIQL where the Q-network is dropped. Quasimetric Reinforcement Learning (\textbf{QRL})~\cite{wang2023optimal} is a goal-conditioned algorithm that fits an asymmetric value function.

\subsection{Finite Scalar Quantization}

We utilize Finite Scalar Quantization (FSQ)~\cite{mentzer2024finite} in the bottleneck of our encoder-decoder neural network architecture to discover discrete skill quantizations. FSQ does not learn an explicit codebook during training as in the VQ-VAE architecture~\cite{van2017neural}, avoiding codebook collapse and training instabilities typical of VQ-VAE. Instead, it utilizes scalar quantization by applying a bounding function $f$ to constrain the $D$-dimensional continuous representation $e\in \mathbb{R}^D$, followed by element-wise rounding to the nearest integer. The bounding function $f$ is chosen as
\begin{equation}
    f(e_d) = \lfloor N_d/2 \rfloor \sigma(e_d)
\end{equation}
to restrict each dimension $d$ of the resulting $D$-dimensional quantized vector $\hat{z} = \text{round}(f(e))$ to take on one of $N_d$ unique integer values. To solve the non-differentiability issue of the rounding operation, FSQ utilizes the Straight-Through Estimator (STE)~\cite{bengio2013estimating} to propagate gradients during training. The quantized vectors $\hat{z}$ in FSQ construct an implicit codebook $\mathcal{Z}$ with a total capacity of $K=\prod_{d=1}^{D} N_d$, which we refer to as the codebook size. Therefore, each quantized vector $\hat{z}$ can be mapped bijectively to an integer index $k\in\{1, \dots, K\}$. The codebook size $K$ is therefore controlled by the choice of a specified set of FSQ layer dimensions $\{N_d\}_{d=1}^{D}$. Throughout the paper, we work directly with these integer indices and denote $\mathcal{Z}$ to be identified with the codebook indices when convenient. 

\section{Effect-Driven Temporal Skill Abstractions}

We aim to learn a goal-conditioned policy from an offline dataset $\mathcal{D} = \{(s_t, a_t, r_{t+1}, s_{t+1})\}$ to solve sparse-reward navigation tasks. Instead of training a policy over the continuous action space, we introduce an intermediate discrete skill space between the policy and the environment, and train the policy over this discrete vocabulary. We learn the discrete skills from data and train a separate low-level controller to translate the chosen skill codes back into executable continuous actions.

The framework, as illustrated in Figure~\ref{fig:stages}, consists of three stages. \textbf{Stage I} (Section~\ref{subsec:stage1}) learns an encoder that maps the action chunks $a_{t:t+L}$ of length $L$ to one of $K$ discrete skill codes $k$ based on their \emph{effect} on the agent's task-relevant features $x$. \textbf{Stage II} (Section~\ref{subsec:stage2}) trains a low-level controller $\pi_\mathcal{A} (k,s_t)$ that translates skill codes into continuous actions. \textbf{Stage III} (Section~\ref{subsec:stage3}) trains a discrete action policy $\pi_{\mathcal{Z}}( s_t, g)$ that selects one of the learned skill codes. We implement discrete variants of four popular methods for learning offline goal-conditioned policies: GCBC, GCIQL, GCIVL, and QRL.

\begin{figure}[!b]
  \centering
  \includegraphics[width=\linewidth]{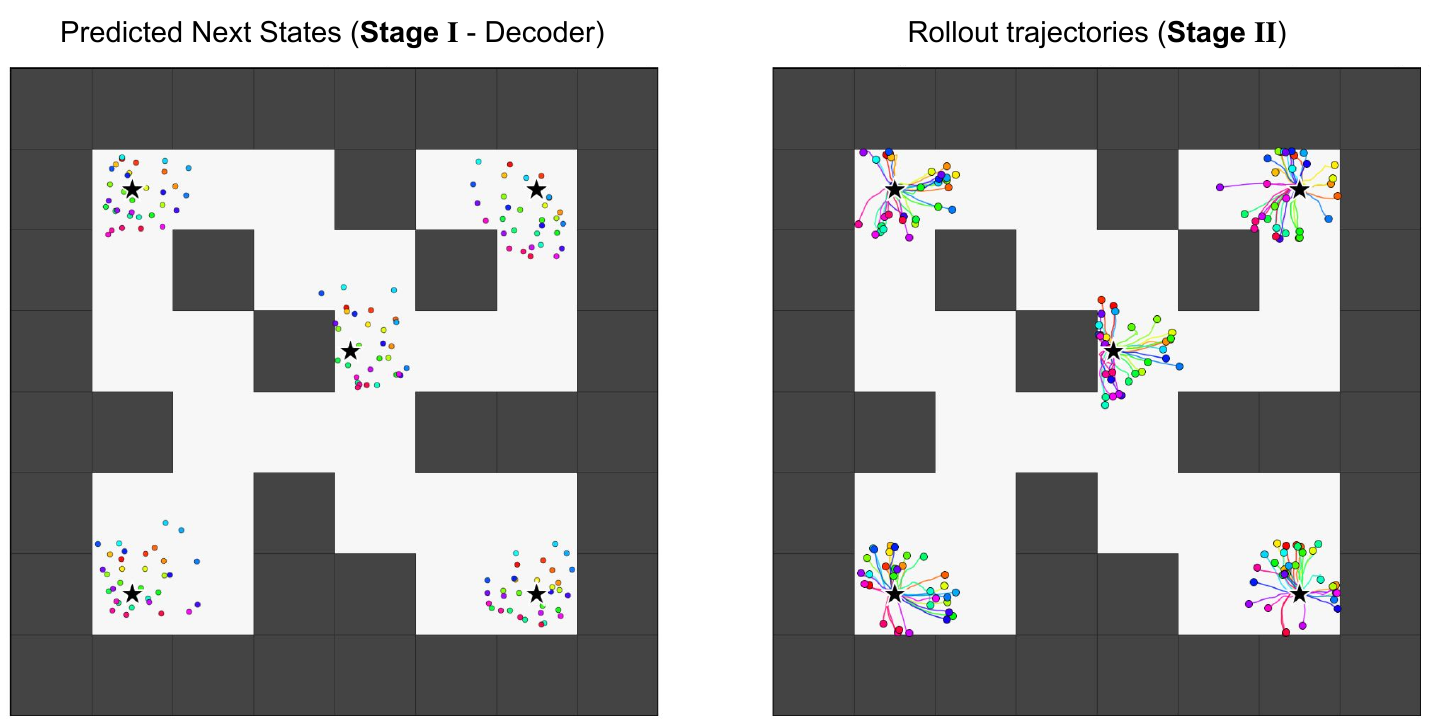}
  \caption{Learned skills in the humanoidmaze-medium-navigate-v0 environment at five different states (black stars). \emph{Left:} next-state predictions by the skills discovered at Stage~I. \emph{Right:} sample rollouts of the low-level controller trained at Stage~II.}
  \label{fig:skills_humanoidmaze}
\end{figure}

\subsection{Stage I: Learning Predictive Skill Codes}\label{subsec:stage1}

The goal of this stage is to learn a vocabulary of discrete skill codes $\{1, \dots, K\}$ from a given offline dataset. We use an encoder-decoder architecture with a discrete bottleneck as shown in Figure~\ref{fig:stages} (Stage~I). 
The encoder $k = \phi(a_{t:t+L})$ maps action chunks $a_{t:t+L}$ sampled from $\mathcal{D}$ to one of the $K$ codes. A chunk first passes through 1-D convolutional layers along the time axis to compress its $L$ actions into a single embedding that summarizes their temporal behavior. Then, the embedding is passed through an FSQ block to discretize it into a skill code $k$. 

The decoder $\hat{x}_{t+L} = \psi(k, s_t)$ takes a skill code $k$, which represents a chunk of $L$ actions, together with the current state $s_t$, to predict $x_{t+L}$, the agent's task-relevant next-state features after the action chunk is executed. Instead of predicting the full next state which may contain high-dimensional variables such as orientation, joint angles, and velocities we predict only the task-relevant next-state features. Predicting the full state would force the discretization bottleneck to encode noisy information irrelevant to the task's success criteria. By limiting the discretization to goal-relevant features, we ensure that each discrete skill is defined by the outcome that matters for the task.

The encoder-decoder architecture is trained to minimize the MSE loss
\begin{equation}
    L_{\text{pred}} = \left\| x_{t+L} - \psi\bigl( \phi(a_{t:t+L}),\, s_t \bigr) \right\|^2_2
\end{equation}
between the actual and predicted future task-relevant features of the agent.

Once trained, we run $\phi$ over the entire dataset to translate every continuous action chunk $a_{t:t+L}$ to a discrete skill code $k$, producing the new discrete transitions dataset $\overline{\mathcal{D}}=\{(s_t,k,s_{t+L})\}$ that will be utilized in Stage~III. The Stage~I architecture is used solely to construct the skill vocabulary, and neither $\phi$ nor $\psi$ will be used during evaluation rollouts in the environment. Figure~\ref{fig:skills_humanoidmaze} (left) illustrates the learned codes and their predicted outcomes for five sample states. Each code represents a distinct directional outcome, and the outcomes depend on the current state.

\subsection{Stage II: Learning Low-Level Controller}\label{subsec:stage2}
To act in the environment, we need to convert the discrete skill codes learned in Stage~I to continuous executable actions. We learn such a mapping by training a low-level controller $\pi_\mathcal{A}(k,s_t )$ that produces single-step continuous actions $a_t$ given a skill code and the current fully-observed state. The controller is implemented as a simple feedforward neural network. The skill code $k\in\{1,\dots, K\}$ is mapped to an embedding, and the state $s_t$ is passed through a state encoder. The two are then concatenated and passed through the decoder to output the action through a linear layer. The network is optimized to minimize the MSE loss
\begin{equation}
    L_{\text{ctrl}} = \left\| a_{t} - \pi_\mathcal{A}(k,s_t ) \right\|^2_2.
\end{equation}

Since $\pi_\mathcal{A}$ only depends on the current states and on the learned skill vocabulary from Stage~I, the same low-level controller is used across all four offline RL methods at Stage~III (Section~\ref{subsec:stage3}). Figure~\ref{fig:skills_humanoidmaze} (right) shows rollouts starting from five sample states in the environment by running $\pi_\mathcal{A}$ with a fixed skill code $k$ iteratively over $L$ time steps. 

\subsection{Stage III: Learning High-Level Discrete Policy}\label{subsec:stage3}
After we transferred the continuous observations to discrete transitions ($\overline{\mathcal{D}}=\{(s_t,k,s_{t+L})\}$) in the previous stages,  we can train a high-level goal-conditioned policy $\pi_{\mathcal{Z}}(k \mid s_t, g)$ where $g$ is a hindsight-relabeled goal~\cite{andrychowicz2017hindsight} sampled from a future state in the trajectory or a random state in the dataset, augmenting the discrete dataset $\overline{\mathcal{D}}$ to $\{(s_t,k,s_{t+L}, g)\}$.
The high-level policy selects a code $k \in \{1,\dots, K\}$ by outputting a categorical distribution over the $K$ skill code indices. To demonstrate that our method is \emph{policy-independent}, we instantiate $\pi_{\mathcal{Z}}$ with discrete variants of four standard offline goal-conditioned RL methods: simple imitation learning (GCBC)~\cite{ghosh2021learning}, implicit value-learning with and without a $Q$ network (GCIQL, GCIVL)~\cite{kostrikov2021offline, park2025ogbench}, and quasimetric RL (QRL)~\cite{wang2023optimal}. All these variants are trained on the same discrete dataset $\overline{\mathcal{D}}$ produced in Stage~I and use the same low-level controller $\pi_\mathcal{A}$ trained in Stage~II for evaluation rollouts. The variants share the same discrete actor architecture, a Multi-Layer Perceptron (MLP) mapping $(s_t,g)$ to $K$ logits. 

Below, we propose discrete extensions of these methods, introducing novel training objectives tailored to operate over a discrete skill space. We refer to these extensions, along with their training objectives, as D-GCBC, D-GCIQL, D-GCIVL, and D-QRL, respectively.

\textbf{D-GCBC.} Unlike the rest of the methods, GCBC does not learn any auxiliary value and is trained to purely imitate demonstrated behavior. The discrete actor is trained to assign high probability to the target skill code $k$ via the cross-entropy loss
\begin{equation}\label{eq:cross-entropy}
    L_{\text{D-GCBC}}
    = -\,\mathbb{E}_{(s_t, k, g) \sim \overline{\mathcal{D}}}\!\left[
       \log \pi_{\mathcal{Z}}(k \mid s_t, g)
   \right].
\end{equation}

\textbf{D-GCIQL.} The discrete variant of goal-conditioned implicit $Q$-learning (GCIQL)~\cite{kostrikov2021offline} utilizes Advantage Weighted Regression (AWR)~\cite{peng2019advantage} to weight the cross-entropy loss (\ref{eq:cross-entropy}) by an advantage estimated from a learned value function. Standard GCIQL learns a critic $Q$ network $Q(s, g, a)$ and a value network $V(s, g)$ fitted to $Q$ via expectile regression~\cite{newey1987asymmetric}. Since the skill space is finite in D-GCIQL, we implement the critic as a discrete network $Q: \mathcal{S} \times \mathcal{G} \to \mathbb{R}^K$ that produces a value per skill code. The critic's Bellman target is the post-skill state $s_{t+L}$. Thus the critic valuates $L$-step skills, minimizing
\begin{equation}
    L^Q_{\text{D-GCIQL}}
    = \mathbb{E}_{\overline{\mathcal{D}}}\!\left[
        \bigl( Q(s_t, g, k) - \bigl[ r(s_{t+L}, g) + \gamma\, V(s_{t+L}, g) \bigr] \bigr)^{\!2}
      \right]
\end{equation}
where $r(s, g) = -1[s \neq g]$ defines the sparse reward function. The actor head's loss is weighted with advantage $A_Q(s_t, g, k) = Q(s_t, g, k) - V(s_t, g)$ by a temperature hyperparameter $\alpha$:
\begin{equation}
    L^{\text{actor}}_{\text{D-GCIQL}}
    = -\,\mathbb{E}_{\overline{\mathcal{D}}}\!\left[
        e^{\alpha A_Q(s_t, g, k)}
        \log \pi_{\mathcal{Z}}(k \mid s_t, g)
      \right]
\end{equation}

\textbf{D-GCIVL.} The standard goal-conditioned implicit $V$-learning (GCIVL)~\cite{park2025ogbench} simplifies the GCIQL architecture by eliminating the $Q$ network and trains only a $V$ network alongside the actor head. In D-GCIVL, the advantage is computed directly from the temporal difference of the value across one $L$-step skill as $A_V(s_t, s_{t+L}, g) = V(s_{t+L}, g) - V(s_t, g)$. The $V$ network is trained on expectile regression~\cite{newey1987asymmetric} on the skill-scale Bellman residual. The actor is trained with the same AWR-weighted cross-entropy loss as GCIQL:
\begin{equation}\label{eq:gcivl_loss}
    L^{\text{actor}}_{\text{D-GCIVL}}
    = -\,\mathbb{E}_{\overline{\mathcal{D}}}\!\left[
        e^{\alpha A_V(s_t, s_{t+L}, g)}
        \log \pi_{\mathcal{Z}}(k \mid s_t, g)
      \right]
\end{equation}

\textbf{D-QRL.} Quasimetric RL (QRL)~\cite{wang2023optimal} computes the goal-conditioned \emph{quasimetric distance} $d(s,g)$ that estimates how many steps away the agent is from reaching the goal $g$. The distance is asymmetric (a quasimetric), so generally $d(s,g)\neq d(g,s)$. It is implemented as an Interval Quasimetric Embedding (IQE)~\cite{tongzhouw2022iqe} that satisfies the triangle inequality. The learned distance acts as a negative value $V(s,g)=-d(s,g)$, and the skill-step advantage is computed as
\begin{equation}\label{eq:qrl-adv}
    A_V(s_t, s_{t+L}, g) \;=\; V(s_{t+L}, g) - V(s_t, g) \;=\; d(s_t, g) - d(s_{t+L}, g).
\end{equation}

Following~\cite{wang2023optimal}, we learn $d$ by pushing random state-goal pairs apart while keeping consecutive steps within unit distance. In D-QRL, we apply this in the skill scale by enforcing $d(s_t, s_{t+L}) \leq 1$. The actor head is trained with the same advantage-weighted cross-entropy loss (\ref{eq:gcivl_loss}).

\section{Experiments}
We conduct our experiments to investigate the following questions:
(1)~Does our outcome-predictive quantization produce more diverse and better-separated skills than alternatives that discretize based on action chunk reconstruction, full proprioceptive state-prediction, or single-step predictions?~(Section~\ref{subsec:stageI_qual})
(2)~Do the resulting skill vocabularies translate into effective downstream task performance?~(Section~\ref{subsec:stageI_quant})
(3)~Are the discrete variants more effective than the continuous-actor versions across various offline RL methods?~(Section~\ref{subsec:full_eval_exp})

\subsection{Experiment Setup}

We evaluate our framework on tasks from the Offline Goal-Conditioned RL Benchmark \textbf{OGBench}~\cite{park2025ogbench}, D4RL Kitchen~\cite{fu2021d4rl}, and the Push-T dataset~\cite{chi2023diffusion}. The OGBench Maze tasks contain three locomotion environments with different control dimensions: 2D point mass (\textbf{pointmaze}), an 8-DoF Ant (\textbf{antmaze}), and a 21-DoF Humanoid (\textbf{humanoidmaze}). We test on three distinct maze types available in the environment: medium, large, and giant. For each agent and maze type, we test our framework on two of the dataset types. The \textbf{navigate} dataset is collected by an expert policy that navigates the agent in the maze in long trajectories. The \textbf{stitch} dataset is composed of short-horizon trajectories, requiring the learned policies to join trajectories to reach a longer-horizon goal, thus testing the ability of an agent to generalize to distant unseen goals~\cite{yang2023essential}.

For state-based tasks, we define the task-related state features $x$ using the variables that define the success criteria in the tasks: the agent's position for Maze, button colors for OGBench Puzzle, the object states for D4RL Kitchen, and the T-Block pose for Push-T task. For pixel-based OGBench navigation and manipulation tasks, Stage~I instead predicts the full future image observation, requiring no selection of task-relevant features.

\subsection{Quality of the Learned Skill Vocabulary}\label{subsec:stageI_qual}
\begin{figure}[h]
  \includegraphics[width=\linewidth]{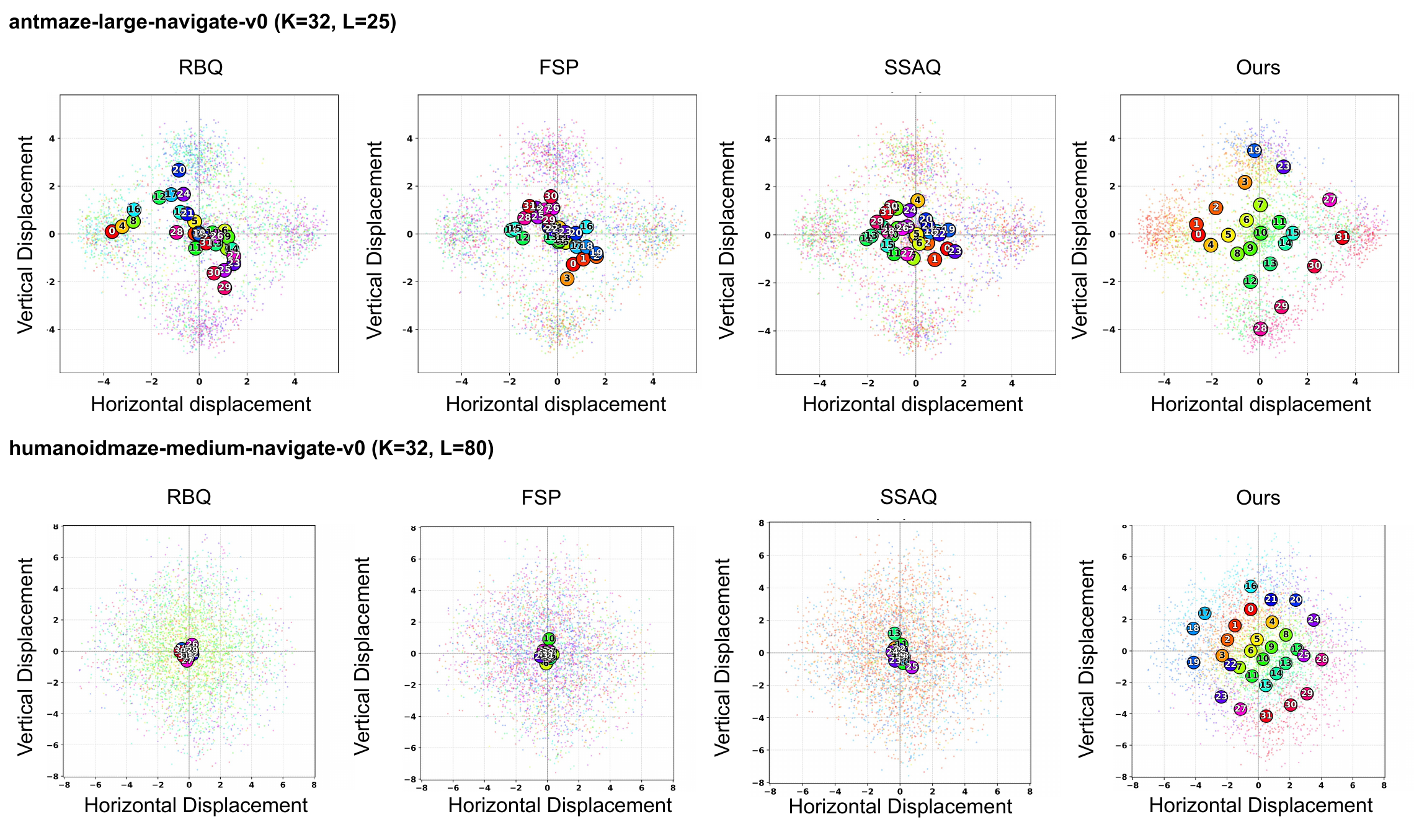}
  \centering
  \caption{Displacement clusters in the offline dataset colored by skill code produced by the four quantization methods in Stage‑I on antmaze-large-navigate (top) and humanoidmaze-medium-navigate (bottom). The baseline methods are Reconstruction-Based Quantization (RBQ), Full-State Prediction (FSP), and Single-Step Action Quantization (SSAQ), as detailed in the text. Each dot represents the displacement entailed by an action chunk $a_{t:t+L}$ sampled from the offline dataset and is colored by the corresponding code $k$. Markers locate the centroids of all action-chunk samples mapped to the same $k$.}
  \label{fig:stage1_ablation_vis}
\end{figure}

We evaluate the effect of three design choices in our Stage~I architecture: (1) predicting only the agent's position vs.\ the entire proprioceptive state, (2) outcome prediction objective vs.\ action chunk reconstruction objective, and (3) quantizing an $L$-length action chunk vs.\ single-step actions. To this end, we implemented three variants of quantization methods in Stage~I, each isolating one of the axes we would like to test: Full-State Prediction (FSP)-based discretization, which quantizes action chunks based on the full proprioceptive state; Reconstruction-Based Quantization (RBQ), which is a state-conditioned autoencoder that reconstructs action chunks, and the Single-Step Action Quantization (SSAQ), which keeps the prediction objective on the agent's position but quantizes single-step actions instead of $L$-length chunks. In every Stage~I configuration, the Stage~II and Stage~III architectures are kept the same to test the different discretization approaches in isolation.

The qualitative comparisons in Figure~\ref{fig:stage1_ablation_vis} show that our outcome-predictive action chunk quantization on agent positions produces the best-separated set of skills, with the code centroids spanning a larger range of displacements in distinct directions. The clear separation is what we want from a useful skill library, so a high-level policy selecting among them can have a meaningful set of choices for reaching different goals. In contrast, the RBQ collapses the centroids near the origin largely due to quantizing the skills based on the low-level action modalities instead of spatial outcomes. FSP fails to obtain spatially distinct skills, since it discretizes the action chunks based on high-frequency and not directly goal-related features such as torso orientation, joint angles, and velocities. By doing so, it wastes the codebook capacity on capturing the low-level movements, destroying the objective of discovering high-level skill abstractions. The SSAQ method fails to produce consistent skill codes over a trajectory. The failures are more severe in higher-dimensional action spaces. On humanoidmaze (Figure~\ref{fig:stage1_ablation_vis} bottom row), almost all of the codes are collapsed near the origin in the baselines since the high-dimensional prediction objective in both the action reconstruction and full next-state prediction methods leads to codes that are uninformative about the goal. In contrast, our method predicts only the goal-relevant features, keeping the quantization objective low-dimensional and thereby less affected by the increasing complexity of the state or actions. 

\subsection{Downstream Performance of Different Quantization Methods}\label{subsec:stageI_quant}
We aim to evaluate the downstream performance of our quantization method in goal-reaching tasks and compare it against existing quantization baselines (FSP, SSAQ, and RBQ). To this end, we used the skill vocabularies constructed in Sec.~\ref {subsec:stageI_qual} for each Stage~I configuration. Then, using the learned skill vocabularies, we trained a low-level controller (Stage~II) and a high-level policy (Stage~III) for each configuration. We chose D-GCBC as the common high-level policy in the state-based tasks due to its simplicity, since it is purely imitation-based with no value-learning, so it didn't require different adaptations for each configuration. We evaluated the methods on various navigate and stitch datasets, and reported the results of evaluation rollouts in Table~\ref{tab:ablation_study_next_state_pred}. Across all eleven evaluated tasks, the results further confirm that the qualitative differences in the learned skill vocabularies translate into downstream navigation performance. 

\begin{table}[b]
    \centering
    \caption{Success Rate ($\%$) for trained policies that use different Stage I architectures: Ours, Full-State Prediction (FSP), Single-Step Action Quantization (SSAQ), and Reconstruction Based Quantization (RBQ). The runs are taken with GCBC for state-based and GCIQL for pixel-based tasks. The results report the mean and standard deviation of the success rates over 5 seeds. Bold values indicate statistically best or indifferent models~(Welch's t-test, p>0.05).}
    \label{tab:ablation_study_next_state_pred}
    
    \begin{tabular}{lllll}
        \toprule
        \textbf{Environment} &  \textbf{Ours} & \textbf{FSP} & \textbf{SSAQ} & \textbf{RBQ}\\
        \midrule
        antmaze-medium-navigate      & $\mathbf{72 \pm 2}$  & $27 \pm 3$  & $10 \pm 1$ & $21 \pm 13$ \\
        antmaze-large-navigate       & $\mathbf{46 \pm 4}$  & $19 \pm 2$  & $9 \pm 2$ & $15 \pm 8$ \\
        antmaze-giant-navigate       & $\mathbf{4 \pm 1}$   & $0 \pm 0$   & $0 \pm 0$ & $0 \pm 0$ \\
        \addlinespace
        antmaze-medium-stitch        & $\mathbf{60 \pm 9}$ & $20 \pm 3$  & $49 \pm 7$ & $\mathbf{53 \pm 8}$ \\
        antmaze-large-stitch         & $\mathbf{16 \pm 2}$  & $10 \pm 0$  & $11 \pm 3$ & $\mathbf{12 \pm 5}$ \\
        antmaze-giant-stitch         & $\mathbf{0 \pm 0}$   & $\mathbf{0 \pm 0}$  & $\mathbf{0 \pm 0}$ & $\mathbf{0 \pm 0}$ \\
        \addlinespace
        humanoidmaze-medium-navigate & $\mathbf{16 \pm 1}$  & $2 \pm 1$   & $2 \pm 1$ & $2 \pm 1$ \\
        humanoidmaze-large-navigate  & $\mathbf{5 \pm 1}$   & $1 \pm 0$   & $1 \pm 0$ & $1 \pm 0$ \\
        humanoidmaze-giant-navigate  & $\mathbf{2 \pm 1}$   & $0 \pm 0$   & $0 \pm 0$ & $0 \pm 0$ \\
        \addlinespace
        visual-antmaze-medium-navigate & $\mathbf{93 \pm 2}$ & $-$ & $26 \pm 6$ & $16 \pm 6$ \\
        visual-antmaze-medium-stitch & $\mathbf{78 \pm 9}$ & $-$ & $6 \pm 1$ & $5 \pm 1$ \\
    \bottomrule
    \end{tabular}

\end{table}

\subsection{Efficacy of Discrete High-Level Policies}\label{subsec:full_eval_exp}
\begin{table}[t!]
  \caption{Performance comparison across various environments. The results are reported as the mean and standard deviations of success rates over 8 seeds for state-based OGBench tasks and D4RL Kitchen, 4 seeds for pixel-based OGBench tasks, and the T-block's maximum coverage of the goal pose in Push-T over 5 seeds. Bold values denote the statistically best or indifferent variant for each high-level policy (Welch's t-test, p>0.05).}
  \label{tab:maze-performance-table}
  \centering
  \small
  \setlength{\tabcolsep}{4pt}
  
  \begin{tabular}{l cccccccc} 
    \toprule
    & \multicolumn{2}{c}{\textbf{GCBC}} & \multicolumn{2}{c}{\textbf{GCIQL}} & \multicolumn{2}{c}{\textbf{GCIVL}} & \multicolumn{2}{c}{\textbf{QRL}} \\
    \cmidrule(lr){2-3} \cmidrule(lr){4-5} \cmidrule(lr){6-7} \cmidrule(lr){8-9} 
    \textbf{Environment} & \textbf{Base} & \textbf{D-} & \textbf{Base} & \textbf{D-} & \textbf{Base} & \textbf{D-} & \textbf{Base} & \textbf{D-} \\
    \midrule
    \multicolumn{9}{l}{\textbf{pointmaze}} \\
    \quad medium-navigate      & $9 \pm 6$  & $\mathbf{37 \pm 7}$  & $53 \pm 8$ & $\mathbf{85 \pm 4}$& $63 \pm 6$ & $\mathbf{89 \pm 3}$ & $82 \pm 5$ & $\mathbf{88 \pm 5}$ \\
    \quad large-navigate      & $29 \pm 6$ & $\mathbf{67 \pm 11}$ & $45 \pm 5$ & $\mathbf{67 \pm 5}$& $34 \pm 5$ & $\mathbf{88 \pm 6}$ & $\mathbf{86 \pm 9}$ & $\mathbf{91 \pm 2}$ \\
    \quad giant-navigate      & $1 \pm 2$  & $\mathbf{7 \pm 6}$   & $0 \pm 0$  & $\mathbf{78 \pm 7}$  & $0 \pm 0$  & $\mathbf{82 \pm 7}$ & $68 \pm 7$ & $\mathbf{75 \pm 1}$  \\
    \addlinespace
    \quad medium-stitch      & $\mathbf{23 \pm 18}$& $\mathbf{35 \pm 12}$ & $21 \pm 9$ & $\mathbf{56 \pm 7}$  & $\mathbf{70 \pm 14}$& $\mathbf{77 \pm 2}$ & $\mathbf{80 \pm 12}$& $\mathbf{87 \pm 2}$  \\
    \quad large-stitch      & $\mathbf{7 \pm 5}$  & $\mathbf{6 \pm 8}$ & $\mathbf{31 \pm 2}$  & $19 \pm 7$  & $12 \pm 6$ & $\mathbf{43 \pm 9}$ & $\mathbf{84 \pm 15}$& $60 \pm 14$   \\
    \quad giant-stitch      & $\mathbf{0 \pm 0}$  & $\mathbf{0 \pm 0}$  & $\mathbf{0 \pm 0}$  & $\mathbf{0 \pm 1}$  & $0 \pm 0$  & $\mathbf{22 \pm 9}$ & $\mathbf{50 \pm 8}$ & $\mathbf{55 \pm 12}$  \\
    \addlinespace
    \multicolumn{9}{l}{\textbf{antmaze}} \\
    \quad medium-navigate      & $29 \pm 4$ & $\mathbf{72 \pm 4}$ & $71 \pm 4$ & $\mathbf{84 \pm 3}$ & $72 \pm 8$ & $\mathbf{88 \pm 2}$ & $\mathbf{88 \pm 3}$ & $\mathbf{90 \pm 2}$  \\
    \quad large-navigate      & $24 \pm 2$ & $\mathbf{45 \pm 4}$ & $34 \pm 4$ & $\mathbf{58 \pm 8}$ & $16 \pm 5$ & $\mathbf{70 \pm 7}$ & $\mathbf{75 \pm 6}$ & $\mathbf{76 \pm 3}$ \\
    \quad giant-navigate      & $0 \pm 0$  & $\mathbf{4 \pm 1}$  & $0 \pm 0$  & $\mathbf{4 \pm 1}$  & $0 \pm 0$  & $\mathbf{7 \pm 2}$  & $\mathbf{14 \pm 3}$ & $\mathbf{13 \pm 4}$  \\
    \addlinespace
    \quad medium-stitch      & $45 \pm 11$& $\mathbf{62 \pm 8}$ & $29 \pm 6$ & $\mathbf{55 \pm 6}$ & $44 \pm 6$ & $\mathbf{64 \pm 4}$ & $59 \pm 7$ & $\mathbf{69 \pm 5}$ \\
    \quad large-stitch      & $3 \pm 3$  & $\mathbf{16 \pm 2}$ & $7 \pm 2$  & $\mathbf{21 \pm 1}$ & $18 \pm 2$ & $\mathbf{29 \pm 4}$ & $18 \pm 2$ & $\mathbf{38 \pm 3}$ \\
    \quad giant-stitch      & $\mathbf{0 \pm 0}$  & $\mathbf{0 \pm 0}$  & $\mathbf{0 \pm 0}$  & $\mathbf{0 \pm 0}$   & $0 \pm 0$  & $\mathbf{1 \pm 0}$  & $0 \pm 0$  & $\mathbf{2 \pm 1}$  \\
    \addlinespace
    \multicolumn{9}{l}{\textbf{humanoidmaze}} \\
    \quad medium-navigate      & $8 \pm 2$  & $\mathbf{15 \pm 1}$ & $27 \pm 2$ & $\mathbf{38 \pm 2}$ & $24 \pm 2$ & $\mathbf{38 \pm 2}$ & $21 \pm 8$ & $\mathbf{53 \pm 3}$ \\
    \quad large-navigate      & $1 \pm 0$  & $\mathbf{5 \pm 1}$  & $2 \pm 1$  & $\mathbf{14 \pm 2}$ & $2 \pm 1$  & $\mathbf{12 \pm 2}$ & $5 \pm 1$  & $\mathbf{22 \pm 6}$  \\
    \quad giant-navigate      & $0 \pm 0$  & $\mathbf{2 \pm 1}$  & $0 \pm 0$  & $\mathbf{1 \pm 0}$ & $0 \pm 0$  & $\mathbf{3 \pm 1}$  & $1 \pm 0$  & $\mathbf{24 \pm 5}$ \\
    \addlinespace
    \quad medium-stitch      & $\mathbf{29 \pm 5}$  & $\mathbf{29 \pm 3}$ & $12 \pm 3$ & $\mathbf{18 \pm 3}$ & $12 \pm 2$ & $\mathbf{21 \pm 4}$ & $18 \pm 2$ & $\mathbf{24 \pm 3}$ \\
    \quad large-stitch      & $\mathbf{6 \pm 3}$  & $\mathbf{6 \pm 3}$  & $0 \pm 0$  & $\mathbf{7 \pm 2}$ & $1 \pm 1$  & $\mathbf{7 \pm 1}$ & $3 \pm 1$  & $\mathbf{6 \pm 2}$  \\
    \quad giant-stitch      & $0 \pm 0$  & $\mathbf{1 \pm 0}$  & $\mathbf{0 \pm 0}$  & $\mathbf{0 \pm 0}$ & $0 \pm 0$  & $\mathbf{1 \pm 0}$  & $0 \pm 0$  & $\mathbf{3 \pm 1}$ \\

    \addlinespace
    \multicolumn{9}{l}{\textbf{manipulation}} \\
    \quad puzzle-4x4-play  & $\mathbf{0 \pm 0}$ & $\mathbf{0 \pm 0}$ & $26 \pm 3$ & $\mathbf{45 \pm 9}$ & $13 \pm 2$ & $\mathbf{18 \pm 5}$ & $0 \pm 0$ & $\mathbf{29 \pm 5}$ \\
    \addlinespace
    \quad kitchen-partial   & $27 \pm 15$  & $\mathbf{52 \pm 11}$  & $40 \pm 9$ & $\mathbf{50 \pm 8}$ & $27 \pm 16$ & $\mathbf{49 \pm 10}$ & $34 \pm 13$ & $\mathbf{47 \pm 10}$   \\
    \quad kitchen-mixed     & $37 \pm 14$ & $\mathbf{61 \pm 5}$  & $39 \pm 15$ & $\mathbf{57 \pm 11}$ & $29 \pm 17$ & $\mathbf{50 \pm 9}$  & $35 \pm 16$ & $\mathbf{55 \pm 9}$   \\
    \quad kitchen-complete  & $13 \pm 6$  & $\mathbf{30 \pm 10}$ & $\mathbf{32 \pm 9}$  & $\mathbf{37 \pm 5}$  & $14 \pm 8$  & $\mathbf{34 \pm 7}$  & $9 \pm 7$   & $\mathbf{32 \pm 7}$\\
    \addlinespace
    \quad Push-T State      & $20 \pm 4$ & $\mathbf{41 \pm 8}$ & $22 \pm 2$ & $\mathbf{37 \pm 10}$ & $25 \pm 2$ & $\mathbf{38 \pm 6}$ & $21 \pm 4$ & $\mathbf{38 \pm 10}$\\

    \addlinespace
    \multicolumn{9}{l}{\textbf{visual-antmaze}} \\
    \quad medium-navigate   & $11 \pm 2$ & $\mathbf{20 \pm 1}$ & $11 \pm 1$ & $\mathbf{93 \pm 1}$  & $22 \pm 2$ & $\mathbf{92 \pm 3}$  & $0 \pm 0$ & $\mathbf{92 \pm 1}$ \\
    \quad large-navigate    & $4 \pm 0$  & $\mathbf{11 \pm 2}$ & $4 \pm 1$  & $\mathbf{69 \pm 6}$  & $5 \pm 1$  & $\mathbf{72 \pm 5}$  & $0 \pm 0$ & $\mathbf{77 \pm 5}$\\
    \quad giant-navigate    & $\mathbf{0 \pm 0}$  & $\mathbf{1 \pm 1}$  & $0 \pm 0$  & $\mathbf{17 \pm 5}$  & $1 \pm 1$  & $\mathbf{14 \pm 2}$  & $0 \pm 0$ & $\mathbf{15 \pm 3}$ \\
    \addlinespace
    \quad medium-stitch     & $\mathbf{67 \pm 4}$ & $\mathbf{70 \pm 8}$ & $2 \pm 0$  & $\mathbf{78 \pm 7}$  & $6 \pm 2$  & $\mathbf{73 \pm 7}$  & $0 \pm 0$ & $\mathbf{73 \pm 8}$\\
    \quad large-stitch      & $\mathbf{24 \pm 3}$ & $15 \pm 5$ & $0 \pm 0$  & $\mathbf{34 \pm 5}$  & $1 \pm 1$  & $\mathbf{44 \pm 4}$  & $1 \pm 1$ & $\mathbf{41 \pm 9}$\\
    \quad giant-stitch      & $0 \pm 0$  & $\mathbf{2 \pm 1}$  & $0 \pm 0$  & $\mathbf{4 \pm 2}$   & $0 \pm 0$  & $\mathbf{8 \pm 3}$   & $0 \pm 0$ & $\mathbf{6 \pm 2}$ \\
    
    \addlinespace
    \multicolumn{9}{l}{\textbf{visual manipulation}} \\
    \quad visual-scene-play      & $12 \pm 2$  & $\mathbf{17 \pm 3}$  & $12 \pm 2$  & $\mathbf{33 \pm 4}$   & $25 \pm 3$  & $\mathbf{34 \pm 4}$   & $10 \pm 1$ & $\mathbf{39 \pm 4}$  \\
    
    \bottomrule
  \end{tabular}

\end{table}

We evaluate the four \emph{D}-variants of our framework, as detailed in Sec.~\ref {subsec:stage3}, against their continuous-actor \emph{Base} versions. Table~\ref{tab:maze-performance-table} reports mean and standard deviation of success rates across three locomotion environments (pointmaze, antmaze, and humanoidmaze), three maze types (medium, large, and giant), and two dataset types (navigate and stitch), totalling $18$ tasks for state-based navigation tasks, $6$ pixel-based navigation tasks in antmaze, 5 state-based manipulation tasks (OGBench Puzzle, D4RL Kitchen, and Push-T), and 1 pixel-based manipulation task (OGBench Visual Scene). For the Push-T task, the T-block's maximum coverage of the goal pose is reported.

The results show that replacing the continuous actor head of GCBC, GCIQL, GCIVL, and QRL with our discrete one significantly improves the performance on the majority of the tasks. Out of 120 comparisons (30 tasks and 4 methods), we improved 97 significantly, 3 regress, and 20 are statistically indistinguishable. This improvement is especially substantial in tasks where the continuous baseline completely fails to compose any goal-reaching behavior. For instance, in pointmaze-giant-navigate, both GCIQL and GCIVL score $0\%$, while their discrete variants reach $78\%$ and $82\%$, respectively. Similarly, in humanoidmaze-giant-navigate, the \emph{longest-horizon, highest-DoF task} in the benchmark suite, requiring 4000 environment steps to reach the goal with a 21-DoF Humanoid, baselines score near-zero while D-QRL achieves a $24\%$ success rate. This pattern shows the difficulty of optimizing continuous actors over long horizons, where a single off-distribution action can change the entire trajectory. Selecting from a small, finite vocabulary of skills mitigates this.

Among the four discrete variants, D-QRL shows the best performance overall; however, D-GCIVL shows the largest relative gains over the continuous version. D-GCBC shows substantial improvements over the continuous GCBC. Since it is the simplest variant among the four, with no value learning at all, it best isolates the contribution of the discretization; any improvement of D-GCBC over the \emph{Base} variant must be due to the discrete skill vocabulary and not to any change to the value-learning mechanism.

Note that, very large navigation tasks appear to still pose insurmountable difficulties to any method. Except for the pointmaze-giant tasks, none of the tested methods achieved high performance on any of the \emph{giant} tasks. Skill learning for long-horizon planning with many degrees of freedom remains an open challenge.

\section{Conclusion}
We present a framework for learning predictive skill abstractions for offline, goal-conditioned reinforcement learning. We demonstrate that learned skills are more effective in performance if they are obtained by action chunk quantization on outcome predictions, rather than by quantization on the reconstruction objective. We show that the learned skill vocabulary and low-level controller are policy-agnostic by pairing them with the discrete variants of four standard offline RL methods: GCBC, GCIVL, GCIQL, and QRL. Our experimental results validate that operating on a small vocabulary of skills rather than the full continuous action space leads to substantial improvements across all four offline RL methods. 

\paragraph{Limitations.} Our quantization method relies on predicting the goal-relevant state features in the state-based tasks. The agent's position in navigation tasks is a natural target. However, in domains where the success criteria are more complex, selecting appropriate target features may not be trivial. Our low-level controller is intentionally kept simple to be able to attribute the gains purely to the discrete skill vocabulary rather than to the used controller. More expressive controllers, such as diffusion policies~\cite{chi2023diffusion} or transformer-based methods~\cite{zhao2023learning}, would likely improve the fidelity of skill execution.

\begin{ack}
This work was supported by the Scientific and Technological Research Council of Turkey (TUBITAK) ARDEB 1001 Program (124E227) and by the INVERSE Project under Grant 101136067, funded by the European Union. We confirm no competing financial interests.
\end{ack}

\bibliographystyle{unsrt}
\bibliography{references}

@inproceedings{kaelbling1993Learning,
  author    = {Leslie Pack Kaelbling},
  title     = {Learning to Achieve Goals},
  booktitle = {Proceedings of the Thirteenth International Joint Conference on Artificial Intelligence (IJCAI)},
  year      = {1993},
  pages     = {1094--1098},
}

@inproceedings{schaul2015universal,
  title = 	 {Universal Value Function Approximators},
  author = 	 {Schaul, Tom and Horgan, Daniel and Gregor, Karol and Silver, David},
  booktitle = 	 {Proceedings of the 32nd International Conference on Machine Learning},
  pages = 	 {1312--1320},
  year = 	 {2015},
  volume = 	 {37},
  series = 	 {Proceedings of Machine Learning Research},
  publisher =    {PMLR},
}

@article{konidaris2019necessity,
title = {On the necessity of abstraction},
journal = {Current Opinion in Behavioral Sciences},
volume = {29},
pages = {1-7},
year = {2019},
note = {Artificial Intelligence},
issn = {2352-1546},
doi = {https://doi.org/10.1016/j.cobeha.2018.11.005},
author = {George Konidaris},
}

@inproceedings{khetarpal2020can,
  title={What can I do here? A theory of affordances in reinforcement learning},
  author={Khetarpal, Khimya and Ahmed, Zafarali and Comanici, Gheorghe and Abel, David and Precup, Doina},
  booktitle={International Conference on Machine Learning},
  pages={5243--5253},
  year={2020},
  organization={PMLR}
}

@article{gibson1977theory,
  title={The theory of affordances},
  author={Gibson, James J},
  journal={Hilldale, USA},
  volume={1},
  number={2},
  pages={67--82},
  year={1977}
}

@article{sutton1999between,
  title={Between MDPs and semi-MDPs: A framework for temporal abstraction in reinforcement learning},
  author={Sutton, Richard S and Precup, Doina and Singh, Satinder},
  journal={Artificial intelligence},
  volume={112},
  number={1-2},
  pages={181--211},
  year={1999},
  publisher={Elsevier}
}

@inproceedings{
eysenbach2019diversity,
title={Diversity is All You Need: Learning Skills without a Reward Function},
author={Benjamin Eysenbach and Abhishek Gupta and Julian Ibarz and Sergey Levine},
booktitle={International Conference on Learning Representations},
year={2019},
}

@inproceedings{
sharma2020dads,
title={Dynamics-Aware Unsupervised Discovery of Skills},
author={Archit Sharma and Shixiang Gu and Sergey Levine and Vikash Kumar and Karol Hausman},
booktitle={International Conference on Learning Representations},
year={2020},
}

@article{andrychowicz2017hindsight,
  title={Hindsight experience replay},
  author={Andrychowicz, Marcin and Wolski, Filip and Ray, Alex and Schneider, Jonas and Fong, Rachel and Welinder, Peter and McGrew, Bob and Tobin, Josh and Abbeel, Pieter and Zaremba, Wojciech},
  journal={Advances in neural information processing systems},
  volume={30},
  year={2017}
}

@inproceedings{li2020generalized,
 author = {Li, Alexander and Pinto, Lerrel and Abbeel, Pieter},
 booktitle = {Advances in Neural Information Processing Systems},
 pages = {7754--7767},
 publisher = {Curran Associates, Inc.},
 title = {Generalized Hindsight for Reinforcement Learning},
 volume = {33},
 year = {2020}
}

@inproceedings{
yang2022rethinking,
title={Rethinking Goal-Conditioned Supervised Learning and Its Connection to Offline {RL}},
author={Rui Yang and Yiming Lu and Wenzhe Li and Hao Sun and Meng Fang and Yali Du and Xiu Li and Lei Han and Chongjie Zhang},
booktitle={International Conference on Learning Representations},
year={2022},
}

@article{mete2024quest,
  title={Quest: Self-supervised skill abstractions for learning continuous control},
  author={Mete, Atharva and Xue, Haotian and Wilcox, Albert and Chen, Yongxin and Garg, Animesh},
  journal={Advances in Neural Information Processing Systems},
  volume={37},
  pages={4062--4089},
  year={2024}
}

@inproceedings{pinto2022bet,
 author = {Shafiullah, Nur Muhammad and Cui, Zichen and Altanzaya, Ariuntuya (Arty) and Pinto, Lerrel},
 booktitle = {Advances in Neural Information Processing Systems},
 pages = {22955--22968},
 publisher = {Curran Associates, Inc.},
 title = {Behavior Transformers: Cloning k modes with one stone},
 volume = {35},
 year = {2022}
}

@inproceedings{lee2024vqbet,
  title = 	 {Behavior Generation with Latent Actions},
  author =       {Lee, Seungjae and Wang, Yibin and Etukuru, Haritheja and Kim, H. Jin and Shafiullah, Nur Muhammad Mahi and Pinto, Lerrel},
  booktitle = 	 {Proceedings of the 41st International Conference on Machine Learning},
  pages = 	 {26991--27008},
  year = 	 {2024},
  volume = 	 {235},
  series = 	 {Proceedings of Machine Learning Research},
  month = 	 {21--27 Jul},
  publisher =    {PMLR}
}

@inproceedings{luo2023action,
  title={Action-quantized offline reinforcement learning for robotic skill learning},
  author={Luo, Jianlan and Dong, Perry and Wu, Jeffrey and Kumar, Aviral and Geng, Xinyang and Levine, Sergey},
  booktitle={Conference on Robot Learning},
  pages={1348--1361},
  year={2023},
  organization={PMLR}
}

@inproceedings{tang2020discretizing,
  title={Discretizing continuous action space for on-policy optimization},
  author={Tang, Yunhao and Agrawal, Shipra},
  booktitle={Proceedings of the aaai conference on artificial intelligence},
  volume={34},
  number={04},
  pages={5981--5988},
  year={2020}
}

@inproceedings{
ahn2025optionaware,
title={Option-aware Temporally Abstracted Value for Offline Goal-Conditioned Reinforcement Learning},
author={Hongjoon Ahn and Heewoong Choi and Jisu Han and Taesup Moon},
booktitle={The Thirty-ninth Annual Conference on Neural Information Processing Systems},
year={2025}
}

@InProceedings{zheng2024prise,
  title = 	 {{PRISE}: {LLM}-Style Sequence Compression for Learning Temporal Action Abstractions in Control},
  author =       {Zheng, Ruijie and Cheng, Ching-An and Daum\'{e} Iii, Hal and Huang, Furong and Kolobov, Andrey},
  booktitle = 	 {Proceedings of the 41st International Conference on Machine Learning},
  pages = 	 {61267--61286},
  year = 	 {2024},
  volume = 	 {235},
  series = 	 {Proceedings of Machine Learning Research},
  month = 	 {21--27 Jul},
  publisher =    {PMLR}
}

@inproceedings{
    jiang2023efficient,
    title={Efficient Planning in a Compact Latent Action Space},
    author={Zhengyao Jiang and Tianjun Zhang and Michael Janner and Yueying Li and Tim Rockt{\"a}schel and Edward Grefenstette and Yuandong Tian},
    booktitle={The Eleventh International Conference on Learning Representations },
    year={2023}
}

@inproceedings{dadashi2022continuous,
  title={Continuous Control with Action Quantization from Demonstrations},
  author={Dadashi, Robert and Hussenot, L{\'e}onard and Vincent, Damien and Girgin, Sertan and Raichuk, Anton and Geist, Matthieu and Pietquin, Olivier},
  booktitle={International Conference on Machine Learning},
  pages={4537--4557},
  year={2022},
  organization={PMLR}
}

@inproceedings{
    ghosh2021learning,
    title={Learning to Reach Goals via Iterated Supervised Learning},
    author={Dibya Ghosh and Abhishek Gupta and Ashwin Reddy and Justin Fu and Coline Manon Devin and Benjamin Eysenbach and Sergey Levine},
    booktitle={International Conference on Learning Representations},
    year={2021},
}

@article{park2023hiql,
  title={Hiql: Offline goal-conditioned rl with latent states as actions},
  author={Park, Seohong and Ghosh, Dibya and Eysenbach, Benjamin and Levine, Sergey},
  journal={Advances in Neural Information Processing Systems},
  volume={36},
  pages={34866--34891},
  year={2023}
}

@inproceedings{kostrikov2021offline,
  title={Offline reinforcement learning with implicit q-learning},
  author={Kostrikov, Ilya and Nair, Ashvin and Levine, Sergey},
  booktitle={International Conference on Learning Representations (ICLR)},
  year={2022}
}

@article{newey1987asymmetric,
  title={Asymmetric least squares estimation and testing},
  author={Newey, Whitney K and Powell, James L},
  journal={Econometrica: Journal of the Econometric Society},
  pages={819--847},
  year={1987},
  publisher={JSTOR}
}

@inproceedings{wang2023optimal,
  title={Optimal goal-reaching reinforcement learning via quasimetric learning},
  author={Wang, Tongzhou and Torralba, Antonio and Isola, Phillip and Zhang, Amy},
  booktitle={International Conference on Machine Learning},
  pages={36411--36430},
  year={2023},
  organization={PMLR}
}

@article{eysenbach2022contrastive,
  title={Contrastive learning as goal-conditioned reinforcement learning},
  author={Eysenbach, Benjamin and Zhang, Tianjun and Levine, Sergey and Salakhutdinov, Russ R},
  journal={Advances in Neural Information Processing Systems},
  volume={35},
  pages={35603--35620},
  year={2022}
}

@inproceedings{
mentzer2024finite,
title={Finite Scalar Quantization: {VQ}-{VAE} Made Simple},
author={Fabian Mentzer and David Minnen and Eirikur Agustsson and Michael Tschannen},
booktitle={The Twelfth International Conference on Learning Representations},
year={2024},
}

@article{van2017neural,
  title={Neural discrete representation learning},
  author={Van Den Oord, Aaron and Vinyals, Oriol and others},
  journal={Advances in neural information processing systems},
  volume={30},
  year={2017}
}

@article{bengio2013estimating,
  title={Estimating or propagating gradients through stochastic neurons for conditional computation},
  author={Bengio, Yoshua and L{\'e}onard, Nicholas and Courville, Aaron},
  journal={arXiv preprint arXiv:1308.3432},
  year={2013}
}

@inproceedings{park2025ogbench,
  title={OGBench: Benchmarking Offline Goal-Conditioned RL},
  author={Park, Seohong and Frans, Kevin and Eysenbach, Benjamin and Levine, Sergey},
  booktitle={International Conference on Learning Representations (ICLR)},
  year={2025},
}

@misc{fu2021d4rl,
      title={D4RL: Datasets for Deep Data-Driven Reinforcement Learning}, 
      author={Justin Fu and Aviral Kumar and Ofir Nachum and George Tucker and Sergey Levine},
      year={2021},
      eprint={2004.07219},
      archivePrefix={arXiv},
      primaryClass={cs.LG}
}

@inproceedings{kingma2015adam,
  title={Adam: A method for stochastic optimization},
  author={Kingma, Diederik P and Ba, Jimmy Lei},
  booktitle={ICLR: international conference on learning representations},
  pages={1--15},
  year={2015}
}

@article{hendrycks2016gaussian,
  title={Gaussian error linear units (gelus)},
  author={Hendrycks, Dan and Gimpel, Kevin},
  journal={arXiv preprint arXiv:1606.08415},
  year={2016}
}

@inproceedings{tongzhouw2022iqe,
  title={Improved Representation of Asymmetrical Distances with Interval Quasimetric Embeddings},
  author={Wang, Tongzhou and Isola, Phillip},
  booktitle={Workshop on Symmetry and Geometry in Neural Representations (NeurReps) at Conference on Neural Information Processing Systems (NeurIPS)},
  note ={Proceedings Track},
  year={2022},
}

@article{levine2020offlinerl,
  title={Offline reinforcement learning: Tutorial, review, and perspectives on open problems},
  author={Levine, Sergey and Kumar, Aviral and Tucker, George and Fu, Justin},
  journal={arXiv preprint arXiv:2005.01643},
  year={2020}
}

@article{peng2019advantage,
  title={Advantage-weighted regression: Simple and scalable off-policy reinforcement learning},
  author={Peng, Xue Bin and Kumar, Aviral and Zhang, Grace and Levine, Sergey},
  journal={arXiv preprint arXiv:1910.00177},
  year={2019}
}

@inproceedings{yang2023essential,
  title={What is essential for unseen goal generalization of offline goal-conditioned rl?},
  author={Yang, Rui and Yong, Lin and Ma, Xiaoteng and Hu, Hao and Zhang, Chongjie and Zhang, Tong},
  booktitle={International Conference on Machine Learning},
  pages={39543--39571},
  year={2023},
  organization={PMLR}
}

@inproceedings{chi2023diffusion, 
    author    = {Cheng Chi AND Siyuan Feng AND Yilun Du AND Zhenjia Xu AND Eric Cousineau AND Benjamin CM Burchfiel AND Shuran Song}, 
    title     = {{Diffusion Policy: Visuomotor Policy Learning via Action Diffusion}}, 
    booktitle = {Proceedings of Robotics: Science and Systems}, 
    year      = {2023}, 
    DOI       = {10.15607/RSS.2023.XIX.026} 
}

@inproceedings{zhao2023learning,
  title     = {Learning Fine-Grained Bimanual Manipulation with Low-Cost Hardware},
  author    = {Zhao, Tony Z. and Kumar, Vikash and Levine, Sergey and Finn, Chelsea},
  booktitle = {Proceedings of Robotics: Science and Systems (RSS)},
  year      = {2023}
}


\appendix

\section{Technical appendices and supplementary material}

\subsection{Environment Details}
\begin{figure}[h]
    \centering
    \includegraphics[width=\linewidth]{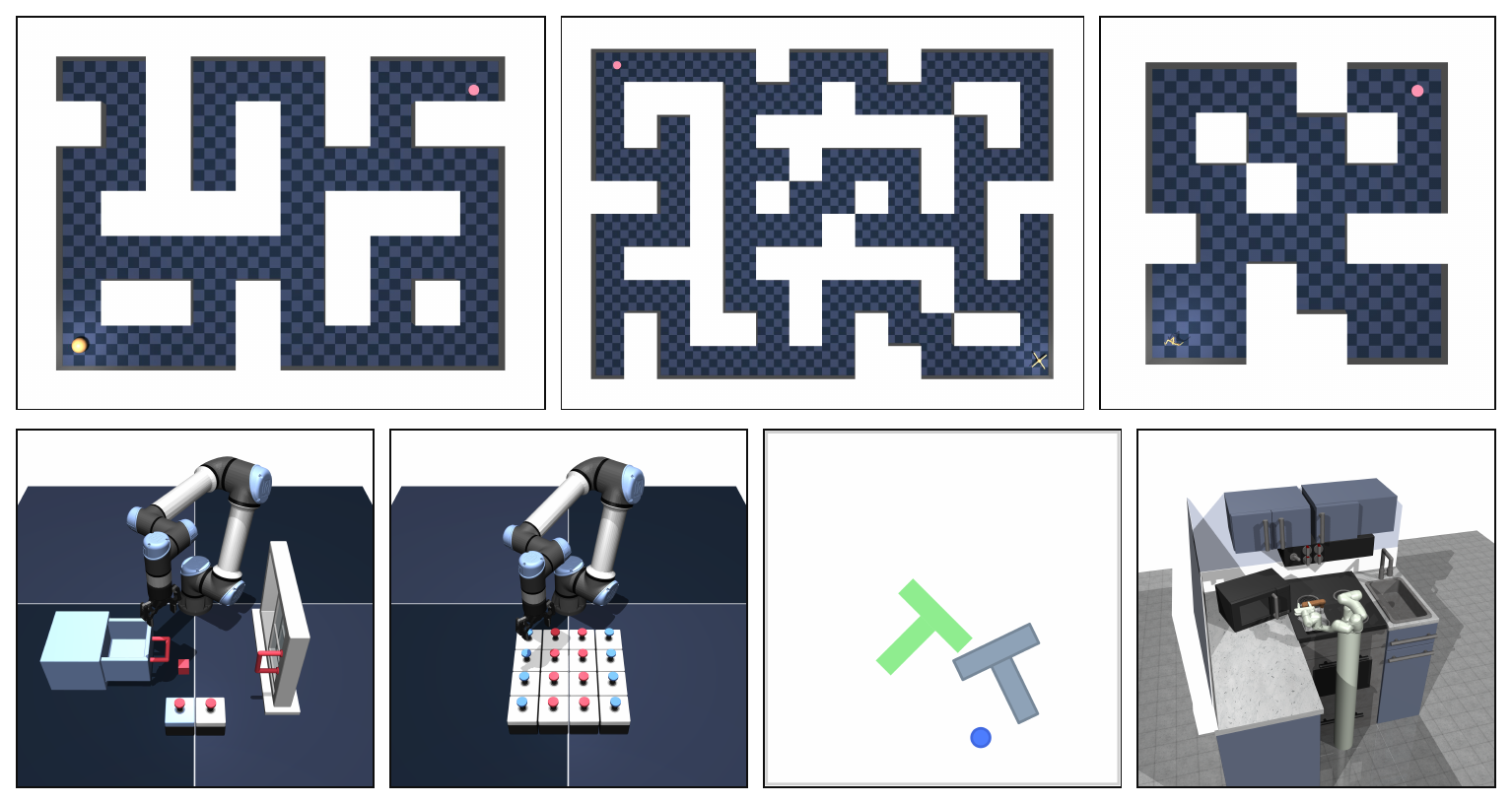}
    \caption{Environments used in our main experiments. (Top) PointMaze large, AntMaze giant, and HumanoidMaze medium~\cite{park2025ogbench}. (Bottom) Scene, Puzzle 4x4~\cite{park2025ogbench}, Push-T~\cite{chi2023diffusion}, and D4RL Kitchen~\cite{fu2021d4rl}.}
    \label{fig:environments}
\end{figure}
The environments used in our experiments are illustrated in Figure~\ref{fig:environments}. We evaluate our method on three locomotion environments from OGBench~\cite{park2025ogbench}: PointMaze, AntMaze, and HumanoidMaze. PointMaze involves controlling a two-dimensional point mass, AntMaze an 8-DoF quadruped, and HumanoidMaze a 21-DoF humanoid. The objective is to navigate the agent to a specified goal location in a maze. We conduct our experiments on three maze sizes (medium, large, and giant) and two dataset types (navigate and stitch). The navigate datasets contain demonstrations of long trajectories, whereas the stitch datasets contain shorter trajectories that require the learned policy to combine skills to perform long-horizon tasks. In Stage I, we use the agent's two-dimensional position on the maze as the prediction target for these environments. We additionally evaluate our approach on Visual AntMaze, where the agent receives $64 \times 64 \times 3$ RGB images instead of proprioceptive observations. The agent must infer its position and orientation from the images and navigate to the specified goal. In the visual tasks, our Stage I predicts the full future image rather than the agent's position. 

We further evaluate our framework on the Puzzle and Visual Scene environments~\cite{park2025ogbench}. In Puzzle-4x4, a robotic arm interacts with a grid of 16 buttons, where pressing a button toggles its color and those of its neighboring buttons. The agent must reach a specified configuration of button colors, requiring combinatorial reasoning and sequential manipulation. We predict the future binary button states in Stage I. Visual Scene involves a robotic arm interacting with multiple objects, including a cube, drawer, window, and buttons controlling the locking mechanisms. The tasks require long sequences of interactions, such as unlocking a drawer and opening it before placing a cube inside it. We use the pixel-based observations and predict the full future image in Stage I. 

Beyond the OGBench~\cite{park2025ogbench} tasks, we also evaluate our framework on Push-T~\cite{chi2023diffusion} and D4RL Kitchen~\cite{fu2021d4rl}. Push-T requires pushing a 2D T-shaped block into a target pose. We use the block's future position and orientation as the prediction target in Stage I. D4RL Kitchen involves a robotic arm manipulating objects in a kitchen environment, including a kettle, a microwave, cabinets, and a light switch. The task is to bring the objects in the scene into a specified goal configuration. We predict the future states of the objects in Stage I.

\subsection{FSQ vs. Vector Quantization}
The discrete bottleneck can be constructed with Vector Quantization (VQ)~\cite{van2017neural} or Finite Scalar Quantization (FSQ)~\cite{mentzer2024finite}. In our method, we preferred FSQ over VQ due to its simple training without auxiliary losses and because it does not suffer from codebook collapse. To compare the approaches, we train our FSQ-based skill quantizer with a VQ-based variant for some of the tasks we used in our experiments. We set the commitment loss coefficient for VQ to $0.25$. Let $k_i$ denote the skill code assigned to the i-th action chunk in the dataset. Given our codebook size $K$ and a dataset of size $|\mathcal{D}|$, let $p_k$ be the empirical probability that code $k$ is selected:
\begin{equation}
    p_k=\frac{1}{|\mathcal{D}|}\sum_{i=1}^{|\mathcal{D}|}{1[k_i=k]}
\end{equation}
The normalized codebook perplexity (NCP) divides the exponential of the entropy of this usage distribution by the codebook size:
\begin{equation}
    \text{NCP} = \frac{1}{K} \exp(-\sum_{k=1}^{K}p_k\text{log} p_k)
\end{equation}
This metric lies between $[\frac{1}{K}, 1]$, where $\frac{1}{K}$ indicates a full codebook collapse and 1 indicates that the codes are used uniformly. The NCP for FSQ and VQ are shown in Table \ref{tab:fsq_vq}. The FSQ-based quantizer consistently reaches higher NCP, indicating better codebook coverage.
\begin{table}[h]
    \centering
    \caption{Normalized Codebook Perplexity for FSQ and VQ (3 seeds)}
    \label{tab:fsq_vq}
    \small
    \setlength{\tabcolsep}{4pt}
    \begin{tabular}{lll}
        \toprule
        \textbf{Environment} &  \textbf{FSQ} & \textbf{VQ} \\
        \midrule
        pointmaze-medium-navigate    & $\mathbf{0.90 \pm 0.02}$   & $0.81 \pm 0.05$ \\
        antmaze-medium-navigate      & $\mathbf{0.79 \pm 0.07}$   & $0.61 \pm 0.05$ \\
        humanoidmaze-medium-navigate & $\mathbf{0.85 \pm 0.01}$   & $0.67 \pm 0.01$ \\
    \bottomrule
    \end{tabular}    
\end{table}

\subsection{Effect of Using a State-Conditioned Encoder}

We argue that mapping action chunks to discrete codes encourages the agent to learn codes that are state-agnostic. For comparison, we create a state-conditioned encoder variant of our Stage~I architecture that adds the state information alongside the action chunk to the FSQ-encoder, leading to quantizations conditioned on both action and state. The results in Table ~\ref{tab:state_conditioning_ablation} show that the state-conditioned quantizer leads to unstable downstream success rates.

\begin{table}[!h]
    \centering
    \caption{Effect of giving state information to the quantizer alongside action chunk on three different maze tasks. We report Success Rate (\%) of the trained GCIQL across 5 seeds.}
    \label{tab:state_conditioning_ablation}
    \small
    \setlength{\tabcolsep}{6pt}
    \begin{tabular}{ll c}
    \toprule
    \textbf{Environment} & \textbf{Ours} &  \textbf{State Conditioned Quantizer}\\
    \midrule
    pointmaze-medium-navigate    & $86 \pm 4$   & $32 \pm 44$\\
    antmaze-medium-navigate      & $83 \pm 2$   & $57 \pm 40$ \\
    humanoidmaze-medium-navigate & $39 \pm 2$   & $10 \pm 17$ \\

    \bottomrule
    \end{tabular}
\end{table}

\subsection{Effect of Codebook Size and Segment Length}
We study the sensitivity of our framework to the codebook size $K$. We trained Stage~I architecture for three different codebook size configurations for each of pointmaze-medium-navigate, antmaze-medium-navigate, and humanoidmaze-medium-navigate. Then, we train Stage~II and D-GCBC as the high-level policy and report the Success Rate in Table~\ref{tab:codebook_size_ablation}. Across all three environments, performance varies at most a few percentages across the test range, indicating that the performance of our framework is insensitive to the choice of hyperparameter $K$. Furthermore, we analyzed the sensitivity of performance to both hyperparameters, $L$ and $K$ on humanoidmaze-medium-navigate, as reported in Table~\ref{tab:codebook_size_chunk_length_ablation}. The results show no significant sensitivity to either $K$ or $L$ in humanoidmaze. The values used in the main experiments were chosen to give the environments a codebook size and segment length comparable to their action dimensionality, but the table suggests that other choices would return comparable results.

\begin{table}[h]
    \centering
    \caption{Effect of codebook size $K$ on the proposed framework. We report 
    Success Rate (\%) of the trained GCBC
    across 5 seeds. Bold indicates the value of $K$ used in the main experiments.}
    \label{tab:codebook_size_ablation}
    \small
    \setlength{\tabcolsep}{6pt}
    \begin{tabular}{ll c}
    \toprule
    \textbf{Environment} & $K$ & \textbf{Success Rate (\%)} \\
    \midrule
    \multirow{3}{*}{pointmaze-medium-navigate}
        & $8$           & $31 \pm 15$ \\
        & $\mathbf{16}$ & $37 \pm 8$ \\
        & $32$          & $38 \pm 13$\\
    \midrule
    \multirow{3}{*}{antmaze-medium-navigate}
        & $16$          & $75 \pm 1$  \\
        & $\mathbf{32}$ & $72 \pm 2$ \\
        & $64$          & $65 \pm 5$ \\
    \midrule
    \multirow{3}{*}{humanoidmaze-medium-navigate}
        & $16$          & $13 \pm 1$ \\
        & $\mathbf{32}$ & $16 \pm 1$ \\
        & $64$          & $17 \pm 2$ \\
    \bottomrule
    \end{tabular}
\end{table}

\begin{table}[h]
    \centering
    \caption{Effect of codebook size $K$ and chunk length $L$ on the humanoidmaze-medium-navigate environment. We report Success Rate (\%) of the trained GCIQL across 5 seeds. Bold indicates the value of configuration used in the main experiments.}
    \label{tab:codebook_size_chunk_length_ablation}
    \small
    \setlength{\tabcolsep}{6pt}
    \begin{tabular}{ll cc}
    \toprule
    \textbf{} & $\mathbf{K=16}$ &  $\mathbf{K=32}$ & $\mathbf{K=64}$ \\
    \midrule
    $\mathbf{L=40}$ & $35 \pm 3$   & $36 \pm 5$ & $37 \pm 2$ \\

    $\mathbf{L=80}$ & $37 \pm 3$   & $\mathbf{39 \pm 2}$ & $40 \pm 4$ \\

    $\mathbf{L=120}$ & $31 \pm 4$  & $35 \pm 2$ & $37 \pm 4$ \\

    \bottomrule
    \end{tabular}
\end{table}

\subsection{Comparisons with Hierarchical Offline RL Methods}

We have compared D-GCIQL and D-QRL with two recent hierarchical offline RL methods, HIQL~\cite{park2023hiql} and OTA~\cite{ahn2025optionaware} on a subset of our main tasks. The results in Table~\ref{tab:comparison_with_hierarchical} show that although HIQL and OTA perform better in AntMaze and HumanoidMaze tasks, the advantage of discrete skill vocabularies on visual maze and puzzle tasks is demonstrated.

\begin{table}[!h]
    \centering
    \caption{We report Success Rate (\%) of the trained D-GCIQL (Ours), D-QRL (Ours), HIQL, and OTA across 8 seeds in state-based and 4 seeds in pixel-based tasks. Bold values indicate the statistically best or indifferent values (Welch's t-test, p>0.05).}
    \label{tab:comparison_with_hierarchical}
    \small
    \setlength{\tabcolsep}{6pt}
    \begin{tabular}{ll ccc}
    \toprule
    \textbf{Environment} & \textbf{D-GCIQL} & \textbf{D-QRL} & \textbf{HIQL} & \textbf{OTA}\\
    \midrule
    pointmaze-medium-navigate        & $\mathbf{85 \pm 4}$ & $\mathbf{88 \pm 5}$ & $79 \pm 5$ & $\mathbf{86 \pm 2}$\\
    pointmaze-large-navigate         & $67 \pm 5$ & $\mathbf{91 \pm 2}$ & $58 \pm 5$ & $85 \pm 5$\\
    antmaze-medium-navigate          & $84 \pm 3$ & $90 \pm 2$ & $\mathbf{96 \pm 1}$ & $\mathbf{96 \pm 1}$\\
    antmaze-medium-stitch            & $55 \pm 6$ & $69 \pm 5$ & $\mathbf{94 \pm 1}$ & $\mathbf{93 \pm 1}$\\
    humanoidmaze-medium-navigate     & $38 \pm 2$ & $53 \pm 3$ & $89 \pm 2$ & $\mathbf{94 \pm 1}$\\
    humanoidmaze-large-stitch        & $7 \pm 2$ & $6 \pm 2$ & $28 \pm 3$ & $\mathbf{57 \pm 3}$\\
    visual-antmaze-large-navigate    & $\mathbf{69 \pm 6}$  & $\mathbf{77 \pm 5}$ & $53 \pm 9$ & $65 \pm 8$\\
    puzzle-4x4-play                  & $\mathbf{45 \pm 9}$  & $29 \pm 5$ & $7 \pm 2$ & $7 \pm 3$\\
    \bottomrule
    \end{tabular}
\end{table}

\subsection{Predicted Effects Visualizations}

\begin{figure}[h]
    \centering
    \includegraphics[width=0.9\linewidth]{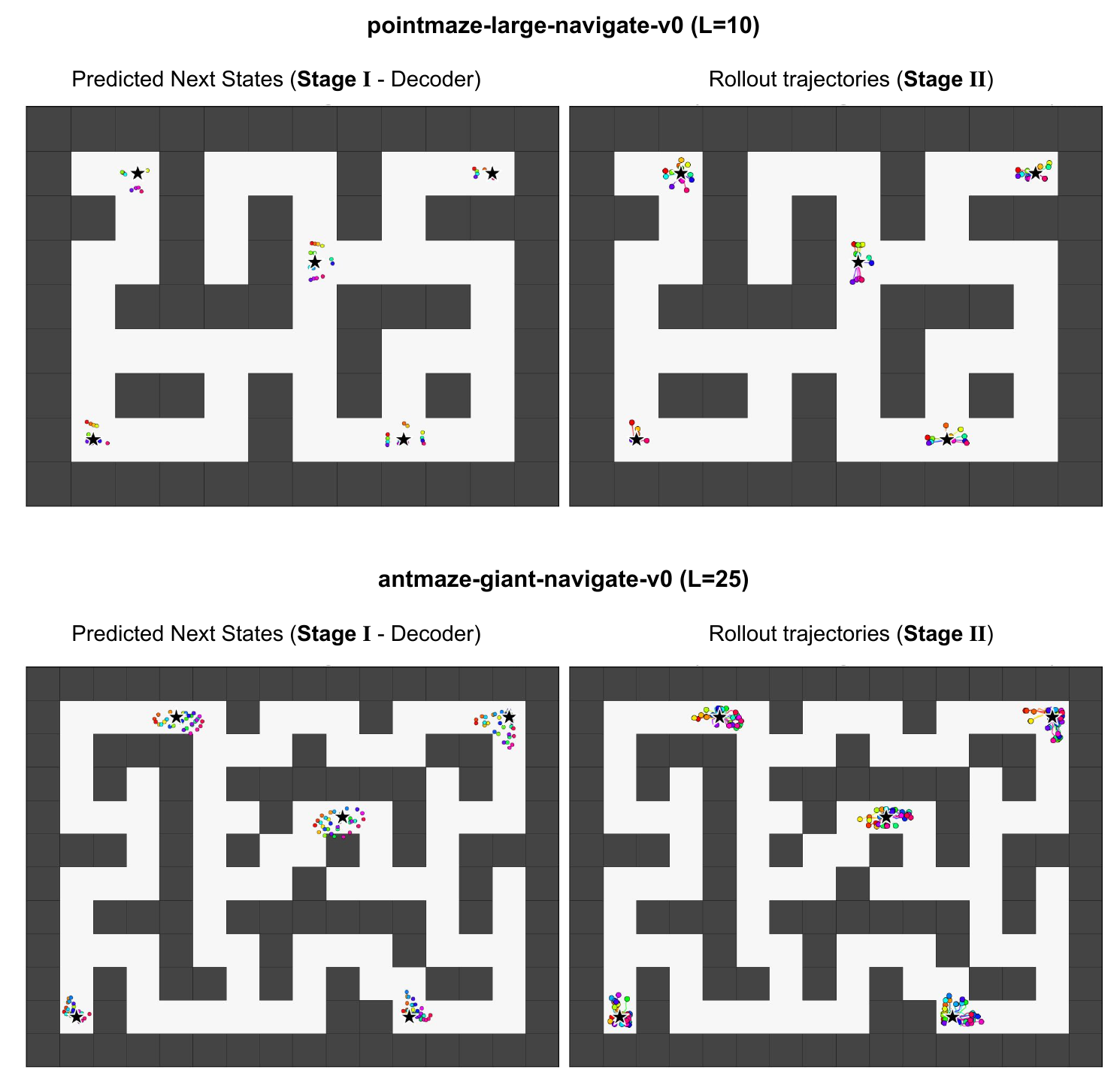}
    \caption{Learned skills in the pointmaze-large-navigate-v0 (top) and the antmaze-giant-navigate-v0 (bottom) environments at five different states (black stars). \emph{Left:} next-state predictions by the skills discovered at Stage~I. \emph{Right:} sample rollouts of the low-level controller trained at Stage~II.}
    \label{fig:extra_stage12}
\end{figure}

Figure~\ref{fig:extra_stage12} shows predicted effects and their executions for PointMaze Large Navigate and AntMaze Giant Navigate tasks. It can be seen that the learned skills are (1) mostly within possibly reachable positions, (2) diverse from each other, (3) executions end up close to the predicted future positions. 

\subsection{Rollout Visualizations with Discrete Skills}

\begin{figure}[!h]
    \centering
    \includegraphics[width=0.9\linewidth]{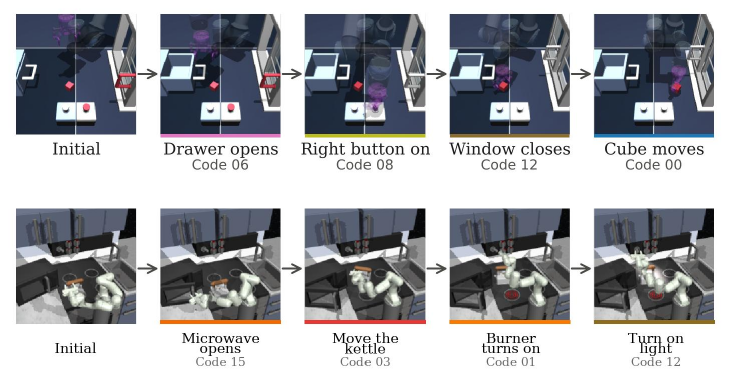}
    \caption{Example of successful rollouts of the discrete (D-) variants on the Visual Scene (top row) and D4RL Kitchen (bottom row) tasks. }
    \label{fig:manip_rollouts}
\end{figure}

\begin{figure}[!h]
    \centering
    \includegraphics[width=\linewidth]{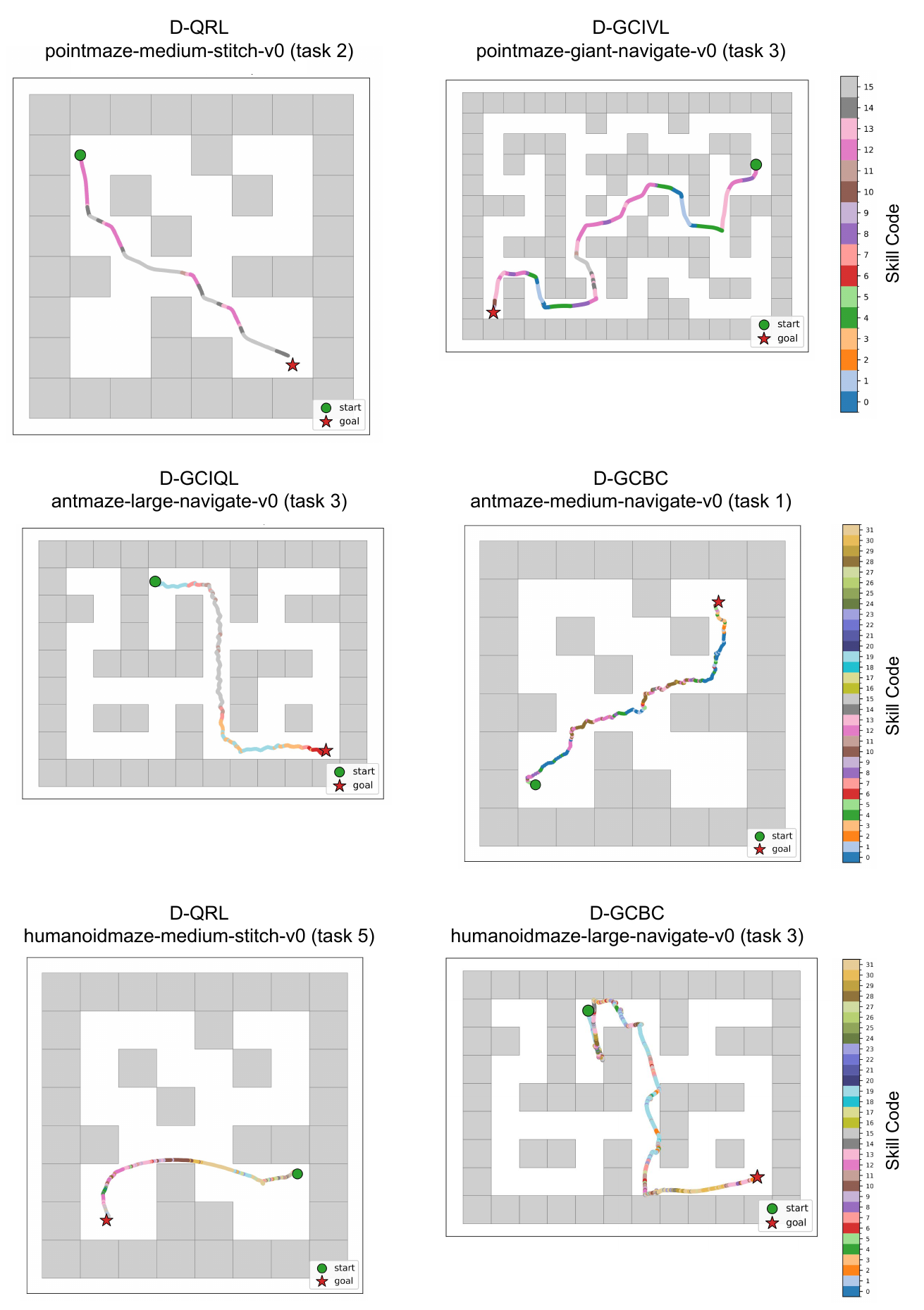}
    \caption{Example of successful rollouts of the discrete (D-) variants on pointmaze, antmaze, and humanoidmaze environments with different dataset types. The colors on the trajectory indicate different skill codes $k\in\{1,\dots,K\}$. Different tasks are the goals defined by the OGBench~\cite{park2025ogbench} benchmark. }
    \label{fig:skill_rollouts}
\end{figure}

Some examples of skill rollouts with D-GCBC, D-GCIQL, D-GCIVL, and D-QRL are shown in Figure~\ref{fig:skill_rollouts} for navigation tasks and in Figure~\ref{fig:manip_rollouts} for manipulation tasks in different datasets. 

\subsection{Implementation Details}
We run our experiments on a workstation with NVIDIA GeForce RTX 4090 (24GB GDDR6X), 128GB RAM, and a 2TB PCIe 4.0 NVMe SSD, on a laptop with NVIDIA GeForce RTX 4080 Mobile (12GB GDDR6), 64GB DDR5-5600 RAM, and a 2TB Samsung PM9A1 PCIe 4.0 NVMe SSD, on 2 cloud GPU pods each equipped with an NVIDIA GeForce RTX 4090 (24GB GDDR6X), AMD EPYC 7B13 64-core CPU (128 threads), 251GB RAM, and ~10.8TB of NVMe SSD storage. Each experiment on a single seed takes between 2 and 4 hours across 3 stages. The total compute time to reproduce our experiments is approximately 2,000 GPU-hours, and additional failed or preliminary runs took approximately 500 GPU-hours. Our code is based on the OGBench~\cite{park2025ogbench} repository, with discrete policy modifications. The evaluations are conducted at each 100K step. For the final results, we take the success rates at the last 3 evaluations (at steps 800K, 900K, 1M) for robust comparisons, following~\cite{park2025ogbench}.

\subsubsection{Hyperparameters}

\begin{table}[h]
\caption{Common hyperparameters.}
\label{tab:hyperparameters}
\centering
\footnotesize
\begin{tabular}{ll}
\toprule
\textbf{Hyperparameter} & \textbf{Value} \\
\midrule
    Learning rate & $0.0003$ \\
    Optimizer & Adam~\cite{kingma2015adam} \\
    \# gradient steps & $1000000$ (Stage 3), $200000$ (Stage 1 and 2) \\
    MLP dimensions & $(512, 512, 512)$ \\
    Nonlinearity & GELU~\cite{hendrycks2016gaussian} \\
    Discount factor $\gamma$ & $0.995$ (\{ant, point\}maze-giant, humanoidmaze), $0.99$ (others) \\
    GCIVL/GCIQL expectile $\kappa$ & $0.9$ \\
    QRL quasimetric & IQE~\cite{tongzhouw2022iqe} \\
    QRL margin $\epsilon$ & $0.05$ \\
    Policy $(p^D_{cur}, p^D_{traj}, p^D_{geom}, p^D_{rand})$ ratio for $p^D_{mixed}$ & $(0, 0.5, 0, 0.5)$ (stitch), $(0, 1, 0, 0)$ (others) \\
    Value $(p^D_{cur}, p^D_{traj}, p^D_{geom}, p^D_{rand})$ ratio for $p^D_{mixed}$ & $(0.2, 0, 0.5, 0.3)$ \\
    AWR temperature $a$ & 3.0 (for D-QRL and D-GCIVL), 10.0 (for D-GCIQL) \\
    Codebook size $K$ & $8$ (Push-T), $16$ (pointmaze, kitchen, scene), \\ 
    & $32$ (antmaze, humanoidmaze, puzzle) \\
    Chunk Length $L$ & $10$ (Push-T, pointmaze), $25$ (antmaze, puzzle, scene), \\
    & $50$ (kitchen), $80$ (humanoidmaze) \\
    FSQ Levels $N$ & $(4, 2)$ ($K=8$) $(4, 4)$ ($K=16$), $(4, 4, 2)$ ($K=32$) \\
    \bottomrule
\end{tabular}
\end{table}


\end{document}